\documentclass{article}
\usepackage{amsmath,graphicx,mlspconf}
\usepackage{xcolor}
\usepackage[numbers,sort&compress]{natbib}
\usepackage{times}
\usepackage{latexsym}
\usepackage[T1]{fontenc}
\usepackage[utf8]{inputenc}
\usepackage{microtype}
\usepackage{inconsolata}
\usepackage{booktabs}
\usepackage{amssymb}
\usepackage{makecell}
\usepackage{tabularx}
\usepackage{url}
\usepackage{algorithm}
\usepackage{algpseudocode}
\usepackage{hyperref} 
\copyrightnotice{979-8-3195-0884-3/26/\$31.00 {\copyright}2026 IEEE}

\toappear{2026 IEEE International Workshop on Machine Learning for Signal Processing, Sep.\ 28-- Oct.\ 1, 2026, Atlanta, USA}

\title{Speculative Macro Commit for Faster Tool-Using Agents}
\name{Anonymous}
\address{Anonymous}

\name{%
    Zeyu Liu$^{\star}$%
    \qquad Souvik Kundu$^{\dagger}$%
    \qquad Peter A. Beerel$^{\star}$%
}

\address{%
    $^{\star}$ University of Southern California, Los Angeles, USA \\%
    $^{\dagger}$ Intel Labs, San Diego, USA%
}

\begin{document}
%\ninept

\maketitle

\begin{abstract}
Tool-using LLM agents spend wall-clock time not only on model inference but also in serial action--observation turns, where each tool call, environment transition, and observation can delay subsequent decisions. 
We introduce \textbf{Speculative Macro Commit} (SMC), a runtime mechanism for a two-tier agent system: a large authoritative actor model produces the official trajectory, while a faster speculative drafter model continuously predicts and executes future action chains on an isolated environment snapshot.
SMC mines recurring multi-action skeletons from training traces and stores them in a macro library used to match against action chains predicted by the drafter at runtime.
When the actor's next tool call matches the first drafted action, SMC commits the remaining pre-executed draft steps, together with their observations, to the official trajectory.
Using Qwen3.5-27B INT4 as the authoritative actor model and Qwen3.5-4B as the speculative drafter model, SMC matches the sequential agent's overall accuracy while reducing latency by 10.23\% over the Speculative Actions (SA) baseline and 18.59\% over sequential execution on the $\tau^2$-Bench Telecom subset. 
On AppWorld, SMC reduces wall time by 7.7\% over SA baseline and 44.9\% over sequential execution, with a small reduction in task completion. 
Overall, SMC provides a practical way to reuse multi-step speculative execution and reduce agent latency beyond single-step speculative actions. Our code is publicly available \href{https://github.com/zeyuliu1037/speculative-macro-commit}{\textcolor{magenta}{here}}.
\end{abstract}

\newcommand{\cem}[1]{\textcolor{blue}{cem: #1}}

\section{Introduction}
\label{sec:introduction}
Language agents are becoming a practical interface to software, data, web, and customer-service workflows~\citep{xu2026evolutiontoolusellm, ling2026agentskillsdatadrivenanalysis,
wei2026agenticreasoninglargelanguage}. 
Unlike a model that makes a single prediction, a tool agent runs a loop: it calls the model to choose an action, executes that action in an environment, observes the result, and conditions its next decision on what it observed~\citep{react-2023}. 
This iterative interaction gives agents much of their flexibility, but it also creates a wall-clock bottleneck that model throughput alone does not capture: a single task can wait through dozens of sequential turns even when the model server is well batched.

This latency has several sources. Repeated large-model calls reprocess long contexts and stress cache management~\citep{kvflow-2025}. Tool orchestration adds scheduling, state, and environment-lifecycle overhead~\citep{thunderagent-2026}. The tools and environments can themselves dominate a request; PASTE reports tool execution at 35\% to 61\% of total latency in representative agents~\citep{paste-2026}. 
Across these sources, the main constraint is sequential dependence: the next prompt usually cannot be built until the previous action returns its observation. 
Token-level speculative decoding~\citep{leviathan-2023-specdec} drafts several tokens with a cheap model and verifies them with the expensive one, but it operates strictly inside one model call and cannot remove this cross-step dependency.

Speculative Actions (SA) lifts the draft-then-verify idea from tokens to actions~\citep{speculative-actions-2026}. 
A fast speculative drafter predicts likely future actions while a slower authoritative actor computes the next verified action; when the actor later agrees, an action that was already launched can be committed. 
However, the core SA mechanism only accepts progress at the granularity of adjacent action steps: a correct guess lets the executor reuse work for the next step, while naively longer lookahead has limited benefits. 
Therefore, how to effectively commit multi-step actions still remains underexplored.
% SA represents a useful demonstration of action-level speculation, but it did not answer the full-runtime question: if a tool agent runs with speculation enabled from start to finish, does the complete task actually finish faster under the same quality metric? This leaves a systems gap between speculative-action opportunity analysis and end-to-end speculative tool-agent execution.
% We build on this actor/speculator framing but target a different engineering problem: the public SA codebase ships self-contained environments for benchmarks such as chess~\citep{speculative-actions-2026} and HotpotQA~\citep{yang2018hotpotqadatasetdiverseexplainable}, whereas long-horizon tool agents need a full runtime. 
A complementary line of work observes that tool-agent traces contain repeated sequences of tool calls. 
AWO~\citep{awo-metatools-2026}, for example, mines recurring tool-call sequences and turns them into deterministic composite meta-tools that bundle multiple actions into one invocation. 
This can remove several reasoning steps when the model chooses the right meta-tool.
% Moreover, long-horizon tool-agent traces contain repeated short sequences of tool calls, so the latency bottleneck is not only the next action but also the decisions that often follow it~\citep{awo-metatools-2026}. 
In our experiments, however, simply exposing mined action sequences as additional callable tools is unreliable: open models like Qwen3.5-27B~\citep{qwen3.5} rarely select these mined macros.
This raises the central question of the paper:
\begin{quote}
\emph{Can recurring action patterns reduce agent latency without requiring the model to choose new meta tools?}
\end{quote}
Our answer is \textbf{Speculative Macro Commit} (SMC), 
where a \emph{macro} is a recurring multi-action pattern mined from execution traces.
% , such as a short sequence of tool calls that often appears after the same kind of state.
A \emph{commit} means accepting already executed draft tool calls and their returned results into the agent's real execution history, so that the large model does not need to choose and execute those steps again. 
% The key reframing is that a mined macro should not be an action chain the model chooses, but guarded runtime state: speculative work that becomes eligible for runtime commit only after the authoritative worker verifies the first drafted action and a state-compatible suffix materializes online. 
% SMC uses macro steps as runtime evidence rather than model-visible tools.
When one macro from the mined macro library is identified in this long action chain, and the first action of the macro matches the authoritative action, SMC attempts a \emph{macro commit}: it commits the remaining matched draft steps and their observations into the official trajectory, skipping the corresponding large-model calls as well as environment delays. 
As a macro commit does not ask the actor to regenerate and verify every skipped step, SMC is approximate rather than strictly lossless but is empirically shown to preserve task quality while reducing latency.

% Because a committed macro skips the authoritative model's forward pass and verification entirely, SMC is not lossless, which is precisely why the guards matter. Concretely, our contributions are:
\begin{enumerate}
    \item \textbf{An end-to-end runtime for action-level speculation.} 
    We build a two-tier runtime in which an authoritative actor model and a speculative drafter model run concurrently throughout complete long-horizon tool-agent tasks.
    The executor measures actual wall-clock latency, lets the drafter run ahead on draft action--observation chains, and falls back to ordinary execution when speculation cannot be used.
    % This runtime measures actual wall-clock latency and supports fallback when speculation fails.
    \item \textbf{Speculative Macro Commit.} 
    We introduce SMC, which shifts mined recurring action patterns from model-side macro selection to executor-side validation.
    The executor first identifies macro matches in the drafter's pre-executed action chain; after the authoritative actor agrees with the first matched draft action, SMC can commit the remaining matched draft steps and their observations into the official trajectory.
    This skips the corresponding large-model calls and avoids waiting again for already returned tool results, making SMC an approximate but guarded latency optimization.

    \item \textbf{Evaluation on full tool-agent benchmarks.} 
    With Qwen3.5-27B INT4 as the authoritative actor and Qwen3.5-4B as the speculative drafter on the $\tau^2$-bench Telecom subset, the SMC run matches the single-model baseline's reward while cutting latency by 18.59\%; and reduces latency by 10.23\% over SA. On AppWorld benchmark, SA reduces the latency by 40.37\% compared with the baseline, and SMC further reduces wall time by 7.64\% with a small accuracy tradeoff. 
\end{enumerate}

\section{Related Work}
\label{sec:background}
\textbf{Speculative decoding and speculative agent execution.}
Speculative decoding accelerates a single model call by letting a cheap drafter model propose tokens that a target model verifies in batches~\cite{leviathan-2023-specdec}. 
This line of work reduces decoding latency inside one model invocation, but a ReAct-style tool agent, which interleaves model reasoning with tool calls over many steps~\cite{react-2023}, also pays latency across invocations, since each tool result must return before the next prompt can be formed. 
Speculative Actions extends the draft-and-verify idea from tokens to agent actions, launching predicted future API calls in parallel with an authoritative actor and committing them only when predictions match~\cite{speculative-actions-2026}. 
Our SMC shares this action-level view but differs in what a commit covers: rather than verifying one action at a time, it commits several already executed draft steps, trading the lossless guarantee of one-step verification for a guarded multi-step approximation.

SMC builds on the same actor--speculator setting but uses a different commit rule.
After the actor verifies the first drafted action, SMC may commit a matched suffix of already executed action--observation pairs without verifying each skipped action with the actor.
This enables multi-step runtime skipping, at the cost of making SMC approximate rather than strictly lossless.

% SMC follows this action-level view, but studies multi-step continuation commit in full tool-agent runtimes where forked environment state, tool observations, and wall-clock serving contention must be handled end to end.

\textbf{Speculative tool execution systems.}
Recent LLM-serving systems attack the same serial LLM--tool loop from complementary angles. 
PASTE speculatively executes likely tool calls while the model is still computing, exploiting stable application-level control flow and data dependencies~\citep{paste-2026}, while ThunderAgent~\citep{thunderagent-2026} and KVFlow~\citep{kvflow-2025} reduce latency through program-aware serving and prefix/cache management without changing which actions enter the trajectory. 
These systems motivate the runtime setting of our work: latency depends on whether useful tool work can move off the serial critical path, not only on model calls. 
% SMC differs by committing a verified multi-action suffix as official trajectory progress, rather than only prefetching, caching, or scheduling individual calls.

\textbf{Workflow compression and meta-tools.}
Agent Workflow Optimization (AWO) mines recurrent tool-call sequences from traces and registers the resulting composites as meta-tools available to the model~\citep{awo-metatools-2026}. 
The key difference is the interface: AWO changes the model-visible toolset and lets the model choose when to call a composite, whereas our SMC keeps the mined steps as hidden runtime state and commits them only after the anchor call is verified.

\section{Speculative Macro Commit}
\label{sec:method}
\subsection{Executor Roles and State}

SMC is implemented in the \emph{agent executor}, the program that runs the tool-agent loop. 
Given the current task history, the executor prompts the model for the next tool call, executes that call, records the returned result, and repeats this process until the task finishes or the step budget is exhausted. The executor is therefore distinct from both the language model and the tool environment: it alone decides which action–result pairs become part of the task history seen by future model calls.

SMC augments this executor with two models that run concurrently. 
The \textbf{authoritative actor} is the large model whose tool calls define the real execution of the agent.  
The \textbf{speculative drafter} is a smaller and faster model that runs ahead of the actor, trying to speculate short future tool-call sequences from the committed history $H$.
The executor maintains $H$, consisting of the tool calls and returned results that future actor prompts will condition on, and a live task state $E$, which is the state reached by executing the calls in $H$. 
The drafter's calls are executed in an isolated draft state $E^i$, so they produce returned results without immediately changing the live task state.
We write the drafter's current output as a draft chain $Q=(q_1,\ldots,q_D)$, where each $q_i$ contains a drafted tool call and the result of executing it in the draft state. 
A \emph{macro} is a recurring multi-action pattern mined from past executions. 
A \emph{commit} is the executor operation that appends executed tool calls and results pairs to $H$ and advances $E$ consistently with them. 
% We use \textbf{commit} only as this runtime operation: append tool-call and observation pairs to $H$, advance $E$ consistently with those pairs, and restart speculation from the updated history. 
When no macro commit is performed, $H$ grows in the usual one-step way: the executor appends the authoritative actor's tool call and its returned result. 
SMC changes this loop through a single addition, the \emph{macro commit rule} of Section~\ref{sec:commit}, which under specific conditions lets one actor step be followed by several already executed draft steps.
% A macro commit changes this update rule only after the actor confirms the first drafted call. 
% Specifically, when the actor's next tool call matches the first drafted call, we call this matched call the \emph{anchor call}. 
% The anchor call itself is committed using the actor's own returned result; only the subsequent drafted calls and results can be accepted from the draft chain. 
% If these subsequent draft steps pass the online checks, the executor appends them to $H$ and advances $E$ accordingly.

% For a macro step, the first drafted call is not copied; it must be regenerated by the actor.
% Only the already executed drafted tool-call and observation pairs that follow that matching first call can be copied into $H$.

% This convention makes the experimental comparison straightforward.
% The same abstraction supports two comparisons: a runtime that runs draft work but disables macro commit, and an SMC runtime that additionally allows multi-step commits. 
When the macro commit rule is disabled, the same executor still runs the actor and drafter concurrently and still reuses single-step speculative work: whenever the actor proposes the same call as the first drafted step, the drafter's already executed call and result are committed directly. 
We call this configuration the \emph{SA-only baseline} and use it as the matched-runtime comparison in Section~\ref{sec:results}.

\subsection{Macro Mining}

% SMC mines macros offline from traces of successful tasks on a training subset, and retained  macros form the \emph{macro library}.
SMC first extracts candidate multi-action patterns from successful training traces.
It then evaluates these candidates using drafter predictions from sampled training states and keeps only those that satisfy the selection criteria below.
The selected candidates form the runtime \emph{macro library} $\mathcal{M}$.
Each macro records an ordered tool-call sequence together with the argument structure needed to match it, replacing task-specific values such as user IDs, document IDs, or timestamps with slots. 
This normalization lets one macro cover repeated behavior across tasks without requiring identical concrete arguments.

% A macro is a short tool-call sequence with reliability metadata. Given trusted trajectories for a task family, we sample states along those trajectories and ask the drafter to propose the next few tool calls. 

Because a macro is useful only if the drafter can reproduce it at deployment, we label candidates under the same actor--drafter setting used at runtime. 
For each sampled training state, the drafter proposes a short future action chain, which we compare against a reference future trajectory from that state. 
A proposal is labeled correct when its tool-call sequence matches the reference future trajectory after task-specific argument values are normalized as described above.
% the mining matcher. 
Each candidate macro thus carries two kinds of evidence: how often the pattern occurs, and how reliably the drafter reproduces it when the relevant state is encountered.

% Each proposed sequence is labeled against the trusted future trajectory, producing exact- and skeleton-level agreement signals. 
% The miner then deduplicates by tool-call skeleton, computes support and a lower confidence bound (LCB) over observed correctness, and emits candidate libraries indexed by support and reliability thresholds. 
% This mining target is intentionally more conservative than raw frequency: a frequent sequence is useful only if it is reliable enough to skip actor calls later.

For a candidate macro $m$, let $n_m$ be the number of labeled opportunities and $k_m$ be the number of correct drafter proposals. 
We compute a conservative reliability estimate using the lower $\delta$-quantile of a Beta posterior,

\begin{equation}
    \underline{p}_m =
F^{-1}_{\mathrm{Beta}(k_m+1,\; n_m-k_m+1)}(\delta)
\end{equation}
and retain $m$ only when $n_m \ge n_{\min}$ and $\underline{p}_m \ge \tau$. 
The thresholds $n_{\min}$ and $\tau$ are selected on held-out tasks using the same executor configuration as the final evaluation.
% so the selected library reflects not only offline frequency but also whether the drafter can produce the pattern early enough to matter.

% Mining is a pruning step, not a safety proof: a library entry only says a pattern is frequent and predictable enough to be considered. 
These criteria are used only to filter the candidate macros.
Whether a retained macro is committed is determined separately at runtime by the macro commit rule.

\subsection{Macro Commit Rule}
\label{sec:commit}

% At runtime, the drafter continuously extends the draft chain $Q$ from the current $H$ and the executor checks whether a suffix of $H$ together with a prefix of $Q$ matches a macro in $\mathcal{M}$.
At runtime, the drafter continuously extends the draft chain $Q$ from the current $H$. Consider a retained macro $m=(u_1,\ldots,u_K)$. A runtime alignment matches $u_1,\ldots,u_j$ to a suffix of $H$ and the remaining $\ell=K-j$ actions to the draft prefix $q_1,\ldots,q_\ell$. The library stores $m$ once; $j$ and $\ell$ are determined by the state reached at runtime.
An alignment is only a commit candidate; it does not change $H$ by itself. The executor waits for the actor's next tool call. If that call matches $q_1$, we call $q_1$ the \emph{anchor call}. The drafter has already executed $q_1$, so the executor commits it together with the drafter's returned result; only $q_2,\ldots,q_\ell$ are committed without individual actor confirmation. This skips $\ell-1$ actor decisions, so candidate selection requires $\ell-1 \ge L_{\min}$. The floor cannot be enforced from $K$ alone: for the same four-action macro, aligning one versus two actions with $H$ leaves three versus two draft actions and therefore skips two versus one actor decisions.

% A match is only a commit candidate; it does not change $H$ by itself. 
% The executor waits for the actor's next tool call. If that call matches the first drafted call $q_1$, we call $q_1$ the \emph{anchor call}. 
% The drafter has already executed $q_1$, so the executor commits $q_1$ together with the drafter's returned result, without executing the call a second time; the subsequent already executed draft steps $q_2,\ldots,q_r$ then become eligible for macro commit. 
% The candidate is accepted only if it skips at least $L_{\min}$ actor decisions, i.e., $r-1 \ge L_{\min}$, since part of the matched macro may already lie in $H$. 
% On acceptance, the executor appends $q_2,\ldots,q_r$ with their returned results to $H$ and advances $E$ accordingly.

Online checks then discard stale draft work and apply benchmark-specific action safety. In AppWorld, manually specified API-name rules reject irreversible or unknown calls, while forkable mutations require a matching live-state replay. If all checks pass, the executor appends $q_2,\ldots,q_\ell$ and their returned results to $H$ and advances $E$ accordingly.
The drafter is restarted only when the committed history actually diverges from the draft chain, that is, when the actor's call does not match $q_1$. 
In that case the executor executes the actor's call in the live state, waits for its result, commits this step, discards the draft work, and restarts the drafter from the updated history. Whenever the actor confirms $q_1$, the executor instead reuses the drafter's execution: it removes the committed steps from $Q$ ($q_1$ alone if no macro commit, $q_1,\ldots,q_r$ if one does), and the drafter keeps extending the remainder of the draft chain, which the executor
continues to search for macro matches in later iterations. 
% If no retained macro matches or any online check fails, only the anchor step is committed, and the executor continues as the SA-only baseline.
% Algorithm~\ref{alg:smc-runtime} summarizes this rule.
If candidate selection finds no eligible alignment (including the depth floor), or if an online check rejects the selected actions, only the anchor step is committed and the executor continues as the SA-only baseline.
Algorithm~\ref{alg:smc-runtime} summarizes this rule.

\begin{algorithm}[t]
\caption{\textsc{Speculative Macro Commit} Executor}
\label{alg:smc-runtime}
\small
\begin{algorithmic}[1]
\Require macro library $\mathcal{M}$, live task state $E$, task $x$
\Require authoritative actor $A_{\mathrm{actor}}$, speculative drafter $A_{\mathrm{draft}}$
\Require draft depth $D$, minimum skipped length $L_{\min}$
\State $H \gets [\,]$ \Comment{committed task history}
\State launch $A_{\mathrm{draft}}$ to maintain a draft chain $Q=(q_1,\ldots,q_D)$ from $H$
\State $F_{\mathrm{actor}} \gets \Call{SubmitActor}{A_{\mathrm{actor}}, H, E}$
\While{task not finished and budget remains}
    \State $a_{\mathrm{actor}} \gets \Call{Wait}{F_{\mathrm{actor}}}$
    \Comment{actor's proposed tool call (not yet executed)}
    \State $Q \gets \Call{LatestDraftChain}{}$
    \If{\Call{FirstActionMatch}{$a_{\mathrm{actor}}, q_1$}}
        \State \Call{Commit}{$H,E,q_1$}
        \Comment{reuse the drafter's executed call and result}
        \State $r \gets 1$
        \State $c \gets \Call{SelectMacro}{\mathcal{M}, H, Q, L_{\min}}$
        \If{$c \ne \bot$}
            \State $(m,r') \gets c$
            \Comment{$m$ matches $q_1,\ldots,q_{r'}$ and $r'-1 \ge L_{\min}$}
            \If{\Call{OnlineChecks}{$m,H,(q_2,\ldots,q_{r'}),x,E$}}
                \For{$q \in (q_2,\ldots,q_{r'})$}
                    \State \Call{Commit}{$H,E,q$}
                \EndFor
                \State $r \gets r'$
            \EndIf
        \EndIf
        \State \Call{PopDraftPrefix}{$Q, r$}
        \Comment{drafter keeps extending the remainder of $Q$}
    \Else
        \State $o \gets \Call{Execute}{a_{\mathrm{actor}}, E}$
        \Comment{run the actor's call in the live state}
        \State \Call{Commit}{$H,E,(a_{\mathrm{actor}}, o)$}
        \State discard draft work; restart $A_{\mathrm{draft}}$ from $H$
    \EndIf
    \State $F_{\mathrm{actor}} \gets \Call{SubmitActor}{A_{\mathrm{actor}}, H, E}$
\EndWhile
\end{algorithmic}
\end{algorithm}

Overall, unlike registered meta-tools, the actor model never emits a macro call: the actor sees the same tool schema as the baseline, and a macro step is a runtime decision over already executed drafter work, not a new action the model must learn. 

\section{EXPERIMENTAL Results}
\label{sec:results}
\subsection{Setup}
\label{sec:setup}
 
\textbf{Models and serving.}
The authoritative actor is a quantized Qwen3.5-27B model, sized to fit on a single GPU; the speculative drafter is Qwen3.5-4B~\citep{qwen3.5}, and both use greedy decoding. 
The sequential baseline runs only the actor on a single GPU. 
The two speculative runtimes use three GPUs: the actor, an actor-class replica that serves speculative requests so they do not queue behind the authoritative request stream, and the drafter. The SA-only runtime (hereafter SA) and SMC share this serving configuration and differ only in whether the macro commit rule of Section~\ref{sec:commit} is enabled.
 
\textbf{Benchmarks and metrics.}
% We evaluate on two full tool-agent benchmarks. $\tau^2$ Telecom contains 2,285 tasks from the Telecom split of $\tau^2$-Bench~\citep{taubench-2024}; AppWorld
% \texttt{test\_normal} contains 168 tasks across 56 scenarios~\citep{appworld-2024}.
We evaluate on two full tool-agent benchmarks, $\tau^2$ Telecom~\citep{taubench-2024} and AppWorld~\citep{appworld-2024}.
% Each task runs under a per-task step cap \texttt{max\_steps}.
We use binary task accuracy for $\tau^2$ Telecom and task-goal completion (TGC) for AppWorld as performance metrics, and average per-task wall time as the latency metric.
% with scenario-goal completion and other diagnostics reported where they clarify boundary behavior. Macro commits and skipped draft steps are reported as mechanism counters, not as success metrics.

\subsection{Main Results}

% Table~\ref{tab:main-results} reports the full-benchmark results on $\tau^2$ Telecom and AppWorld. 
% We separate two effects. The first is the resource-for-latency effect of running speculative workers, measured by the gap between the sequential baseline and speculative-action-only execution (SA). 
% The second is the equal-hardware effect of SMC, measured by the additional gap between SA and SMC under the same 3-GPU speculative setting.

% Table~\ref{tab:main-results} reports full-benchmark results on $\tau^2$ Telecom and AppWorld. 
% We separate two effects: the resource-for-latency effect of adding speculative workers, measured by the gap between the sequential baseline and the SA-only runtime (hereafter SA), and the equal-hardware effect of macro commit, measured by the additional gap between SA and SMC under the same 3-GPU
% speculative setting.

As shown in Table~\ref{tab:main-results} , on $\tau^2$ Telecom, SA reduces average latency from 27.60s to 25.03s per task, 9.31\% below the sequential baseline, and SMC further reduces it to 22.47s, 18.59\% below the baseline and 10.23\% below SA. 
The paired outcomes are unchanged: SMC agrees with the sequential baseline on every task (2,274 correct and 11 incorrect under both runs), and relative to SA it fixes one task and regresses none. 
On this benchmark, macro commit converts committed draft steps into latency savings without changing any task outcome.

AppWorld shows the same direction in a harder regime. 
SA preserves TGC while reducing latency by 40.37\% relative to the sequential baseline, and SMC reaches to 195.9s per task, 44.93\% below the baseline and 7.64\% below SA. 
Unlike $\tau^2$-Bench, this additional speedup comes with a small TGC drop, from 70/168 to 68/168 tasks. 
% We therefore treat AppWorld as a latency-quality boundary case rather than as a lossless acceleration result.

\begin{table}[t]
\centering
\small
\caption{Full benchmark results. Latency is average seconds per task; $\Delta$ is relative to the sequential baseline within each benchmark. $\tau^2$-Bench accuracy is binary task accuracy; AppWorld accuracy is TGC.}
\label{tab:main-results}
\begin{tabular}{llrrr}
\toprule
Benchmark & Run & Acc. & Lat. & $\Delta$(\%) \\
\midrule
\multicolumn{1}{l}{\textit{$\tau^2$ Telecom}}
 & Baseline & 99.52 & 27.60 & -- \\
 & SA & 99.47 & 25.03 & -9.31 \\
 & \textbf{SMC} & \textbf{99.52} & \textbf{22.47} & \textbf{-18.59} \\
\midrule
\multicolumn{1}{l}{\textit{AppWorld}} 
 & Baseline & 41.67 & 355.7 & -- \\
 & SA & 41.67 & 212.1 & -40.37 \\
 & \textbf{SMC} & 40.48 & \textbf{195.9} & \textbf{-44.93} \\
\bottomrule
\end{tabular}
\end{table}

\subsection{Analysis of the AppWorld Gain}
The smaller AppWorld gain over SA is not explained by a lack of macro opportunities. 
Table~\ref{tab:macro-coverage} reports skip density, defined as the number of skipped steps divided by total number of steps in the benchmark.
On AppWorld, SMC commits at least one macro on 62.0\% of tasks and successfully skips 512 actor steps, yielding a skip density of 3.81\%, close to the 3.91\% observed on $\tau^2$ Telecom.
This similarity indicates that, after normalizing for differences in dataset size and step budget, SMC commits a comparable amount of skipped work on both benchmarks.
The two benchmarks therefore differ not in whether SMC finds reusable patterns, but in how reliably the saved work survives into end-to-end wall time.

\begin{table}[t]
\centering
\small
\caption{Macro-step coverage in the main benchmark runs. Skip density divides skipped steps by \texttt{tasks} $\times$ \texttt{max\_steps}.}
\label{tab:macro-coverage}
\begin{tabular}{lrrrr}
\toprule
Benchmark & Commit rate & Hits & Skipped & Skip density \\
\midrule
AppWorld & 62.0\% & 219 & 512 & 3.81\% \\
$\tau^2$ Telecom & 86.2\% & 3,352 & 7,154 & 3.91\% \\
\bottomrule
\end{tabular}
\end{table}

AppWorld is the harder case because task completion, retries, and the per-task step cap (Section~\ref{sec:setup}) all interact with latency. 
An all-task average therefore mixes three effects: skipped draft steps on trajectories that proceed as before, trajectories whose outcome changes, and steps that consume budget without issuing a tool call.
Table~\ref{tab:appworld-slices} controls for the latter two. 
On the 143 tasks where SA and SMC reach the same correctness outcome, SMC reduces total wall time by 13.5\%, substantially more than the 7.64\% all-task gain. 
On the 85 tasks where neither run emits a no-tool-call (NTC) step, the gain is 10.7\%. 
This slice removes idle, recovery-heavy trajectories whose wasted steps inflate wall time and dilute the macro saving It does not hide a side effect of SMC, as SMC produces fewer such steps than SA (68 NTC steps over 31 tasks, versus 105 over 78 for SA), so macro commit reduces rather than induces idle recovery. 
On comparable trajectories, macro commit thus converts skipped draft steps into latency reduction more effectively than the aggregate number suggests; the all-task figure is diluted by outcome shifts and by the idle trajectories isolated above.

\begin{table}[t]
\centering
\small
\caption{AppWorld controlled wall-time slices.}
\label{tab:appworld-slices}
\begin{tabular}{lrrr}
\toprule
Subset & SA & SMC & $\Delta$(\%) \\
\midrule
Same-acc. & 28,920s & 25,014s & -13.5 \\
NTC-free & 16,375s & 14,625s & -10.7 \\
\bottomrule
\end{tabular}
\end{table}

% Macro coverage further supports this interpretation. 
% As shown in Table~\ref{tab:macro-coverage}, AppWorld still fires macros on 62.0\% of tasks and skips 512 suffix actions, with a skip budget close to that of $\tau^2$ Telecom. 
% The difference between the two benchmarks is therefore not whether SMC finds reusable continuations, but whether those continuations remain stable enough for the saved work to appear as clean end-to-end speedup.

% Together, these results delineate the boundary of the method. 
% SMC is most effective when mined patterns are frequent, stable, and outcome preserving, as on $\tau^2$ Telecom. 
% On AppWorld, macro opportunities remain frequent, but the environment is less stable: small trajectory changes can affect completion and correction behavior. 
% Consistent with the guarded-approximation framing of Section~\ref{sec:commit}, AppWorld is evidence that SMC reduces latency on a harder agent benchmark, not evidence of lossless acceleration.

\section{Mechanism Ablations}
\label{sec:ablations}
\subsection{Ablations}
The main results suggest that a macro commit helps only when it satisfies the three conditions built into the commit rule of Section~\ref{sec:commit}. 
The reused steps must stay hidden from the model, since asking the model to select mined patterns adds a decision problem rather than skipping any actor call; the runtime must filter mined candidates aggressively, since a library match alone weakly predicts that the trajectory will follow the mined pattern; and a commit must skip enough actor calls on the critical path to overcome speculation, verification, and synchronization overhead. 
The three ablations below test these conditions in turn.

\subsection{Interface: hidden runtime state}
 
Table~\ref{tab:single-gpu-alternatives} evaluates two simpler ways of using mined routines on a single GPU. 
The first exposes mined patterns as registered meta-tools~\citep{awo-metatools-2026}, making reuse a model-visible action. 
The second leaves the model interface unchanged but commits matched patterns passively, without the online checks of the full commit rule.

Neither alternative has desirable results. 
Registered meta-tools slightly increase latency and lower accuracy, because the actor almost never selects them: exposing mined routines as ordinary tools does not reliably skip actor calls, it poses an extra tool-selection problem. 
Passive committing has the opposite failure mode. It cuts latency by 11.34\%, confirming that hidden runtime reuse can skip actor calls, but accuracy falls from 99.52\% to 96.48\%. 
Hidden reuse alone is therefore insufficient: the runtime must also verify that the committed steps are anchored to the authoritative trajectory. SMC keeps the latency benefit of hidden reuse while adding the online commit conditions.

\begin{table}[t]
\centering
\small
\caption{Results on $\tau^2$-Bench with single GPU setting. }
\label{tab:single-gpu-alternatives}
\begin{tabular}{lrrr}
\toprule
Run & Acc. & Lat. & $\Delta$(\%) \\
\midrule
Baseline & 99.52 & 27.60 & -- \\
Baseline + AWO-like & 99.34 & 27.89 & +1.05 \\
Baseline + passive & 96.48 & 24.47 & -11.34 \\
\bottomrule
\end{tabular}
\end{table}

\subsection{Commit Precision}

The passive-committing result raises the central safety question: how does SMC avoid committing incorrect mined patterns? Table~\ref{tab:filter-precision} answers with a staged audit on 100 held-out $\tau^2$ Telecom tasks.
% not used for mining. 
For the first four rows, SMC suppresses the commit and lets the actor continue, so the realized trajectory provides an event-level counterfactual for whether the candidate steps were exact; the final row reports task-outcome preservation for the steps SMC actually committed.

The mined library by itself is far too noisy to act as a commit rule: a library match reproduces the exact future steps in only 34.6\% of events. 
Requiring the drafter to have already executed the steps raises precision to 70.6\%, since SMC no longer commits patterns that exist only as offline matches. 
Verifying the anchor call raises it to 87.9\% by tying the candidate to the authoritative trajectory, and the depth guard ($L_{\min}=1$, at least one committed step after the anchor) raises it to 90.4\% while removing shallow commits unlikely to repay synchronization overhead. 
Under full rule, all committed events preserve the outcome in this audit. 
As noted in Section~\ref{sec:commit}, this is not a proof that every committed pattern is uniquely correct; it shows that the online checks turn a low-precision library into a high-precision set of commits on this held-out audit.

\begin{table}[t]
\centering
\small
\caption{Conditional precision of the commit stages on 100 held-out $\tau^2$ Telecom tasks. Rows above the rule are event-level exact-match checks; the committed row is a task-outcome check.}
\label{tab:filter-precision}
\begin{tabular}{lrr}
\toprule
Filter stage & Events & \makecell{Correct / \\outcome-preserving} \\
\midrule
Library match only & 1,968 & 681/1,968 (34.6\%) \\
+ drafter-executed & 885 & 625/885 (70.6\%) \\
+ verified anchor call & 711 & 625/711 (87.9\%) \\
+ depth guard ($L_{\min}=1$) & 343 & 310/343 (90.4\%) \\
\midrule
Actually fired & 158 & 158/158 (100.0\%) \\
\bottomrule
\end{tabular}
\end{table}

\subsection{Critical-Path Depth}

High precision alone does not guarantee speedup: a correct macro is unhelpful if it skips only one or two actor calls or skips calls that are off the critical path.
Table~\ref{tab:legacy-final} compares the final SMC runtime with an earlier runtime that committed more aggressively.
The legacy runtime commits almost twice as often as the final runtime and skips more total steps, yet it is 1.64\% slower than SA. 
Raw macro count is thus the wrong objective: many additional legacy commits are too shallow or poorly aligned with already executed draft work, adding commit overhead without reducing end-to-end latency. 
The final runtime suppresses 6,281 depth-one opportunities, enforces the $L_{\min}$ depth floor, and commits only steps already executed on the draft branch. 
While it commits fewer macros and skips fewer total steps, it is 10.23\% faster than SA.
The useful unit is therefore not a macro hit but a guarded, sufficiently deep commit that skips actor calls on the critical path.

\begin{table}[t]
\centering
\small
\caption{$\tau^2$ Telecom full-run comparison: raw macro count versus critical-path commits. Hits is the number of macro commits; Skip is the total committed draft steps after anchors.}
\label{tab:legacy-final}
\begin{tabular}{lrrrrr}
\toprule
Run & Acc. & Lat. & Hits & Skip & $\Delta$(\%) \\
\midrule
SA & 99.47 & 25.03 & 0 & 0 & -- \\
Legacy macro & 99.56 & 25.44 & 6,410 & 10,528 & +1.64 \\
Final SMC & 99.52 & 22.47 & 3,352 & 7,154 & -10.23 \\
\bottomrule
\end{tabular}
\end{table}

\section{Conclusion}
\label{sec:conclusion}

We presented Speculative Macro Commit (SMC), a runtime mechanism that extends speculative action execution from single-step reuse to committing several already executed draft steps after a verified anchor call. 
Because a macro commit is approximate rather than losslessly verified, SMC guards it with conservative offline mining, a verified anchor call, a minimum skipped depth, and online state and argument checks. 
On $\tau^2$ Telecom, SMC gives a quality-preserving latency gain over the full dataset, beyond both the sequential baseline and an equal-hardware speculation-only baseline (SA); on AppWorld it gives a larger wall-time gain with a small completion tradeoff, and larger gains on same-accuracy slices. 
More broadly, workflow macros can live as hidden runtime state rather than model-visible tools, but they help only when the pattern is predictable enough to mine, the drafter has already executed it in the current run, and the commit survives the online checks.

% References should be produced using the bibtex program from suitable
% BiBTeX files (here: strings, refs, manuals). The IEEEbib.bst bibliography
% style file from IEEE produces unsorted bibliography list.
% -------------------------------------------------------------------------
\bibliographystyle{IEEEbib}
\bibliography{strings,refs}

\end{document}